\documentclass[sigconf]{acmart}
\AtBeginDocument{%
  }

\copyrightyear{2026}
\acmYear{2026}
\setcopyright{cc}
\setcctype{by}
\acmConference[MM '26]{Proceedings of the 34th ACM International Conference on Multimedia}{November 10--14, 2026}{Rio de Janeiro, Brazil}
\acmBooktitle{Proceedings of the 34th ACM International Conference on Multimedia (MM '26), November 10--14, 2026, Rio de Janeiro, Brazil}
\acmDOI{10.1145/3767308.3835816}
\acmISBN{979-8-4007-2213-4/2026/11}
\usepackage{algorithm}
\usepackage{algorithmic}
\usepackage{pifont}
\usepackage{multirow}
\usepackage{colortbl}
\newcommand{\cmark}{\ding{51}}
\newcommand{\xmark}{\ding{55}}
\definecolor{lightgray}{gray}{0.9}

\begin{document}

\title{ActFER: Agentic Facial Expression Recognition via Active Tool-Augmented Visual Reasoning}

\author{Shifeng Liu}
\email{lsf0619@mail.ustc.edu.cn}
\affiliation{%
  \institution{State Key Laboratory of Cognitive Intelligence}
  \institution{University of Science and Technology of China}
  \city{Hefei}
  \country{China}}

\author{Zhengye Zhang}
\email{zhengyezhang@mail.ustc.edu.cn}
\affiliation{%
  \institution{State Key Laboratory of Cognitive Intelligence}
  \institution{University of Science and Technology of China}
  \city{Hefei}
  \country{China}}

\author{Sirui Zhao}
\correspondingauthor
\author{Xinglong Mao}
\email{sirui@mail.ustc.edu.cn}
\email{maoxl@mail.ustc.edu.cn}
\affiliation{%
  \institution{State Key Laboratory of Cognitive Intelligence}
  \institution{University of Science and Technology of China}
  \city{Hefei}
  \country{China}}


\author{Zhehan Kan}
\email{kzh24@mails.tsinghua.edu.cn}
\affiliation{%
  \institution{Tsinghua University}
  \city{Beijing}
  \country{China}}

\author{Zhixiang Wei}
\email{zhixiangwei@mail.ustc.edu.cn}
\affiliation{%
  \institution{University of Science and Technology of China}
  \city{Hefei}
  \country{China}}

\author{Shiwei Wu}
\email{davidwu16@sz.tsinghua.edu.cn}
\affiliation{%
  \institution{Tsinghua University}
  \city{Shenzhen}
  \country{China}}

\author{Chaoyou Fu}
\email{bradyfu24@gmail.com}
\affiliation{%
  \institution{Nanjing University}
  \city{Nanjing}
  \country{China}}

\author{Tong Xu}
\email{tongxu@ustc.edu.cn}
\affiliation{%
  \institution{State Key Laboratory of Cognitive Intelligence}
  \institution{University of Science and Technology of China}
  \city{Hefei}
  \country{China}}

\author{Enhong Chen}
\correspondingauthor
\email{cheneh@ustc.edu.cn}
\affiliation{%
  \institution{State Key Laboratory of Cognitive Intelligence}
  \institution{University of Science and Technology of China}
  \city{Hefei}
  \country{China}}

\renewcommand{\shortauthors}{Shifeng Liu et al.}

\begin{abstract}
Recent advances in Multimodal Large Language Models (MLLMs) have created new opportunities for facial expression recognition (FER), moving it beyond pure label prediction toward reasoning-based affect understanding. However, existing MLLM-based FER methods still follow a passive paradigm: they rely on externally prepared facial inputs and perform single-pass reasoning over fixed visual evidence, without the capability for active facial perception. To address this limitation, we propose ActFER, an agentic framework that reformulates FER as active visual evidence acquisition followed by multimodal reasoning. Specifically, ActFER dynamically invokes tools for face detection and alignment, selectively zooms into informative local regions, and reasons over facial Action Units (AUs) and emotions through a visual Chain-of-Thought. To realize such behavior, we further develop Utility-Calibrated GRPO (UC-GRPO), a reinforcement learning algorithm tailored to agentic FER. UC-GRPO uses AU-grounded multi-level verifiable rewards to densify supervision, query-conditional contrastive utility estimation to enable sample-aware dynamic credit assignment for local inspection, and emotion-aware EMA calibration to reduce noisy utility estimates while capturing emotion-wise inspection tendencies. This algorithm enables ActFER to learn both when local inspection is beneficial and how to reason over the acquired evidence. Comprehensive experiments show that ActFER trained with UC-GRPO consistently outperforms passive MLLM-based FER baselines and substantially improves AU prediction accuracy.
\end{abstract}

\ccsdesc[500]{Computing methodologies~Computer vision}
\ccsdesc[300]{Computing methodologies~Reinforcement learning}
\ccsdesc[100]{Human-centered computing~Human computer interaction (HCI)}

\keywords{Facial Expression Recognition, Multimodal Large Language Models, Agentic Reinforcement Learning, Affective Computing}

\maketitle


\section{Introduction}

Multimodal Large Language Models (MLLMs) have shown strong capabilities in visual understanding and cross-modal reasoning~\cite{bai2025qwen3,zhu2025internvl3,openai2025gpt5card,google2025gemini25procard,yin2024nwae403}, creating new opportunities for facial expression recognition (FER), a core task in affective computing and human-centered interaction~\cite{shi2020human,chattopadhyay2020facial}. 
Unlike conventional image classification, FER requires inferring human affect from facial behavior. Although overall facial patterns are informative, accurate recognition often depends on fine-grained local facial movements. The Facial Action Coding System (FACS)~\cite{ekman1978facial} interprets facial muscle activations using Action Units (AUs). Different emotions are often associated with different AU configurations, and FER therefore requires linking local facial evidence to higher-level affective interpretation~\cite{martinez2017automatic}. This makes MLLMs a promising paradigm for moving beyond black-box label prediction toward reasoning-driven affect understanding~\cite{lian2024gpt,yang2024emollm,chen2024finecliper}.

\begin{figure}[t]
    \centering
    \includegraphics[width=\linewidth]{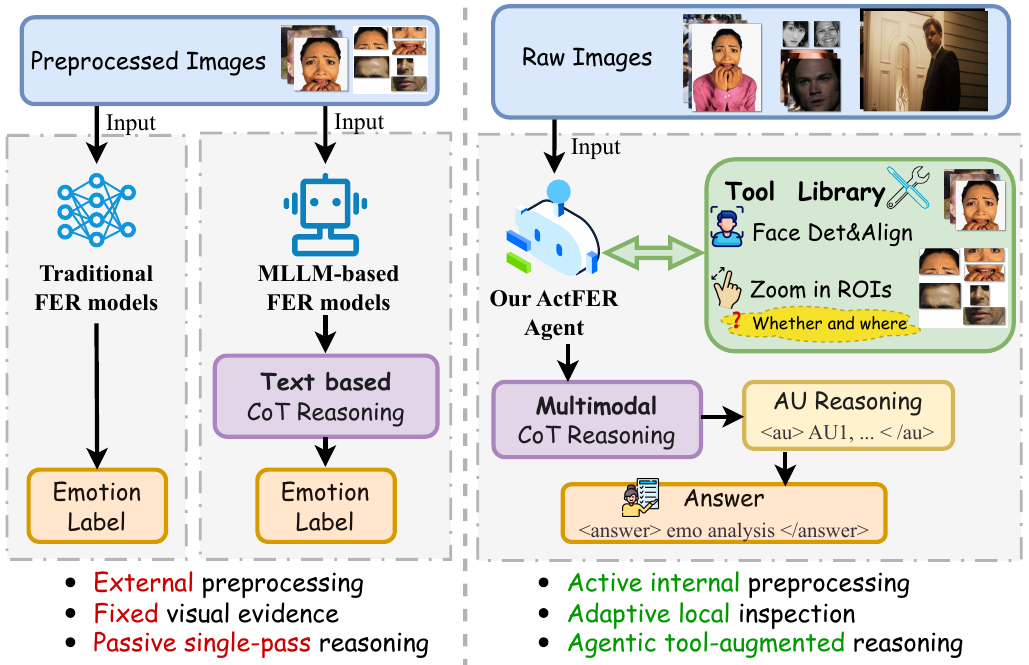}
    \caption{Comparison of FER paradigms. Previous methods rely on passive reasoning over fixed inputs, while ActFER performs tool-augmented agentic inspection.
    }
    \Description{A comparison figure contrasting three paradigms for facial expression recognition. On the left, traditional FER models consume preprocessed face images and output emotion labels. In the middle, MLLM-based FER models take preprocessed faces and perform text-based chain-of-thought reasoning before predicting emotions, reflecting external preprocessing, fixed visual evidence, and single-pass passive reasoning. On the right, ActFER starts from raw images, invokes a tool library for face detection, alignment, and adaptive ROI zooming with an explicit decision of whether and where to zoom, performs multimodal CoT reasoning with AU analysis, and then outputs the final emotion answer, reflecting active internal preprocessing, adaptive local inspection, and tool-augmented agentic reasoning.}
    \label{fig:teaser}
\end{figure}

Recent studies have adapted MLLMs to FER through instruction tuning~\cite{li2024facial,xing2024emo,cheng2024emotion,chaubey2026face}, Chain-of-Thought (CoT) prompting~\cite{Lan2025ExpLLM}, and reinforcement learning (RL) with verifiable rewards~\cite{wu2026facial,zhang2025rethinking}. Despite this progress, existing MLLM-based FER methods still focus on enhancing textual reasoning and treat visual perception as fixed. As illustrated in Figure~\ref{fig:teaser}, they typically rely on externally prepared inputs, such as pre-cropped faces or manually specified local regions, and perform single-pass reasoning over the resulting visual evidence. This passive formulation is restrictive for FER. It leaves the model dependent on preprocessing, weakens robustness to raw in-the-wild inputs, and cannot actively determine when and where additional local inspection is worthwhile.

To overcome this limitation, we propose ActFER, an agentic framework that reformulates FER as active visual evidence acquisition followed by AU-grounded multimodal reasoning. Such a reformulation not only fully leverages the strong language-based reasoning capabilities of MLLMs, but also equips them with the ability to inspect relevant facial evidence, identify meaningful facial movements, and infer emotion from them. Specifically, starting from a raw image, ActFER invokes tools for face detection and alignment to obtain a standardized facial view, then selectively calls a zoom-in tool to inspect informative local regions when finer evidence is needed. This design allows the model to prepare its own analyzable facial evidence instead of relying entirely on external preprocessing, while alignment also provides stable landmark-based spatial references for local inspection. Based on the acquired evidence, ActFER performs structured reasoning over AUs and emotions via a visual CoT, making AU perception an interpretable bridge from local facial movements to final emotion recognition.

Realizing such agentic behavior is challenging because local inspection is not uniformly useful. It can be crucial for complicated or easily confused expressions, but unnecessary for clear samples, and even harmful under poor image quality or unstable alignment. Therefore, the key learning problem is not simply to maximize final emotion accuracy, but to learn whether local inspection is worth performing for the current sample, whether it truly improves affect-relevant facial evidence, and when additional observation should give way to concise reasoning. To this end, we develop Utility-Calibrated GRPO (UC-GRPO), an RL algorithm built upon Group Relative Policy Optimization (GRPO)~\cite{shao2024deepseekmath} and tailored to agentic FER. UC-GRPO addresses this challenge through three key designs. First, AU-grounded multi-level verifiable rewards densify supervision beyond terminal emotion labels and provide intermediate feedback on whether the model has captured meaningful local facial evidence. Second, query-conditional contrastive utility estimation compares zoomed and non-zoomed trajectories within the same rollout group,  enabling sample-aware dynamic credit assignment for local inspection. Third, emotion-aware EMA calibration stabilizes noisy query-conditional utility estimates while capturing emotion-wise inspection tendencies. Together, these components allow ActFER to learn both when local inspection is beneficial and how to reason over the acquired facial evidence.

Overall, our contributions are summarized as follows:

\begin{itemize}
    \item We propose ActFER, a novel agentic framework that reformulates MLLM-based FER as active visual evidence acquisition followed by AU-grounded multimodal reasoning. Starting from raw images, ActFER dynamically invokes tools for face detection, alignment, and adaptive local inspection before final emotion prediction.
    \item To train ActFER effectively, we develop UC-GRPO, an RL algorithm tailored to agentic FER. By combining AU-grounded dense rewards, query-conditional contrastive utility estimation, and emotion-aware EMA calibration, UC-GRPO enables ActFER to learn both when local inspection is worthwhile and how to reason over the acquired facial evidence. 
    \item Extensive experiments demonstrate that ActFER consistently outperforms passive MLLM-based FER baselines and substantially improves AU prediction accuracy. These results further confirm the effectiveness of agentic local inspection and multimodal reasoning for FER. 
\end{itemize}

\begin{figure*}[t]
    \centering
    \includegraphics[width=0.92\textwidth]{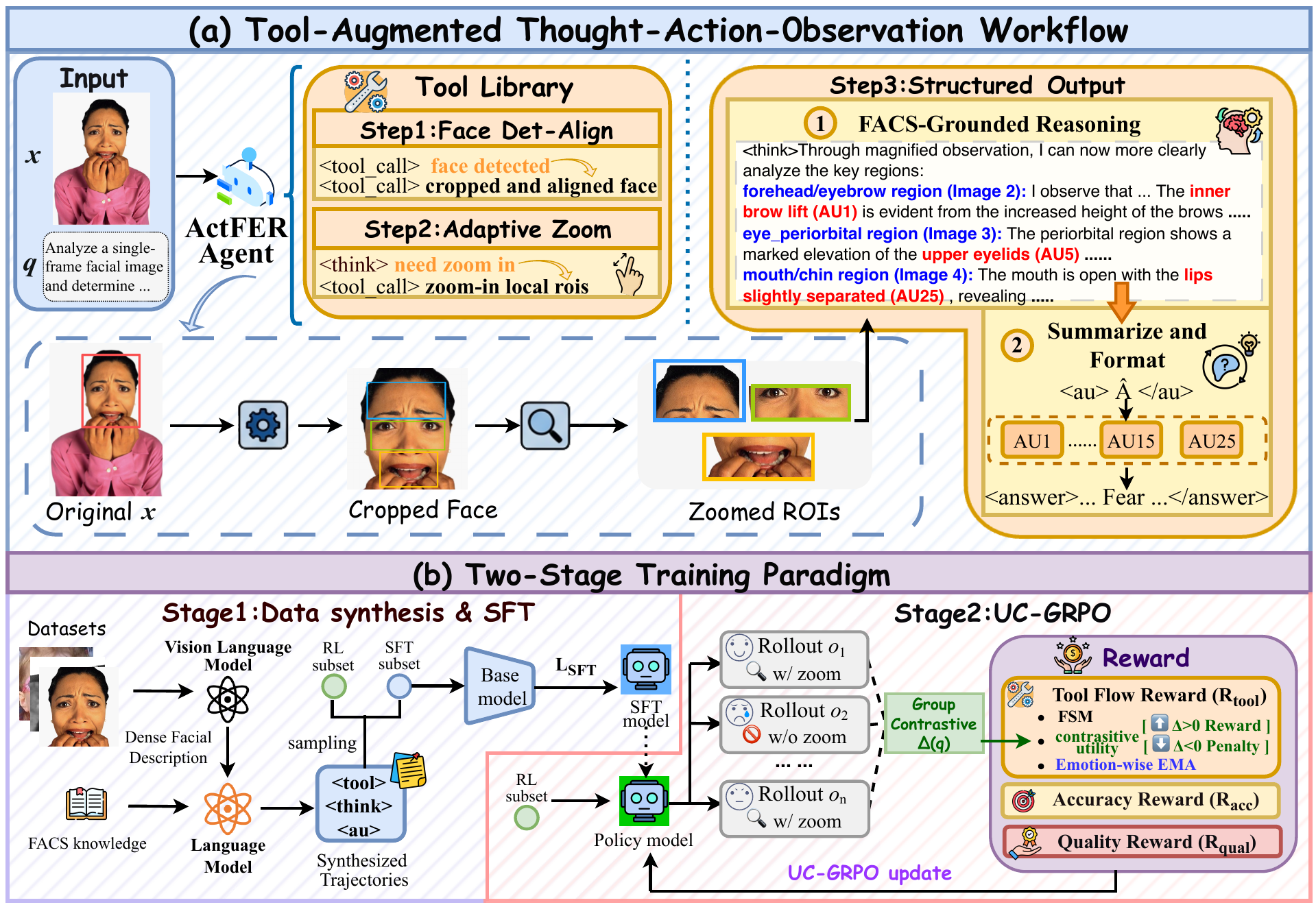}
    \vspace{-0.3cm}
    \caption{Overall architecture of ActFER. The agent combines tool-augmented visual reasoning, perceptual tools, FACS-grounded hierarchical inference, and two-stage SFT+UC-GRPO optimization.}
    \Description{Overall ActFER agent showing the input image, perceptual tool layer, FACS-grounded reasoning process, and two-stage training pipeline.}
    \label{fig:framework_overview}
    \vspace{-0.2cm}
\end{figure*}

\section{Related Work}
\label{sec:related}

\subsection{MLLMs for Facial Expression Recognition}
Traditional FER methods generally cast the task as a multi-class classification problem, focusing on feature extraction and category prediction with deep learning models~\cite{yang2020adaptive,jin2022avt,wu2023patch,sun2023maedfer,li2023intensity,yuan2024auformer,liu2024norface,chen2024static,wang2025qcs,zhao2021twostage,zhao2022meplan,zhao2023dfme,liu2025merclip}. With the rise of MLLMs, recent work has begun to recast FER as a multimodal reasoning problem rather than pure classification. FABA~\cite{li2024facial} introduces instruction tuning and specialized datasets to adapt MLLMs for emotion and AU recognition. ExpLLM~\cite{Lan2025ExpLLM} integrates CoT reasoning into FER, explicitly incorporating AU information into its reasoning templates as interpretable intermediate cues. FEALLM~\cite{hu2025feallm} constructs a dataset aligning facial emotions with AU descriptions to strengthen local-detail modeling. Building on these supervised approaches, recent studies have introduced RL to further refine reasoning paths. UniFER~\cite{zhang2025rethinking} utilizes RL with Verifiable Rewards to enhance FER post-training, while Facial-R1~\cite{wu2026facial} explicitly optimizes the reasoning process using reward signals derived from joint emotion and AU supervision.
Despite these advances, existing methods still lack the capability of active perception, such as dynamically deciding when and where to acquire additional local evidence. ActFER addresses this limitation by reformulating FER as agentic active facial inspection followed by multimodal reasoning.

\subsection{Tool-Augmented Multimodal Reinforcement Learning}

A growing body of work equips large language models with external tools and trains them via RL to decide when and how to invoke those tools during reasoning.
Early efforts such as Toolformer~\cite{schick2023toolformer} show that language models can learn to insert API calls autonomously when doing so improves next-token prediction. Subsequent systems like ToolkenGPT~\cite{hao2023toolkengpt} and ART~\cite{paranjape2023art} extend this idea by representing tool invocations as special tokens or retrieval-augmented templates that can be end-to-end optimized.
In parallel, vision-language agents have begun to adopt RL for multi-step tool use in visual tasks. For example, CogAgent~\cite{hong2024cogagent} and VisualWebArena~\cite{koh2024visualwebarena} train multimodal agents to navigate web interfaces through sequential actions, while Search-o1~\cite{li2025search} and R1-Searcher~\cite{song2025r1} use outcome-driven RL to teach reasoning models when to invoke a retrieval tool and when to rely on internal knowledge.
Despite this progress, existing tool-augmented RL methods mainly target general-purpose tasks and assess tool use only by final task success, making them less suitable for FER, where local zoom-in may be helpful, unnecessary, or even harmful. ActFER addresses this issue with a utility-calibrated policy that learns whether local inspection is worthwhile for each sample.

\section{Method}
\label{sec:method}
\subsection{Overall Agentic Pipeline}

We formulate FER as an agentic, evidence-seeking process rather than a one-shot mapping from images to emotion labels. Given an input face image \(x\) and a query \(q\), the model interacts with the environment through an iterative thought--action--observation loop, progressively acquiring visual evidence and refining its prediction. The episode starts from the initial context \(o_1=\{x,q\}\), where \(o_t\) denotes the interaction context available at step \(t\), consisting of the input and the accumulated action--observation history. At step \(t\), the policy model generates a thought
\begin{equation}
r_t=\pi_{\text{think}}(o_t),
\end{equation}
which serves as the current reasoning state over the available context. This thought either triggers a tool action
\begin{equation}
c_t=\pi_{\text{act}}(o_t,r_t), \qquad c_t \in \mathcal{C},
\end{equation}
or directly terminates the episode with a structured prediction
\begin{equation}
z=(r_t,\hat{A},\hat{y}),
\end{equation}
where \(\hat{A} \subseteq \mathcal{A}\) denotes the predicted AU set and \(\hat{y}\in\mathcal{Y}\) denotes the final emotion label. If an action is executed, the environment returns an observation
\begin{equation}
\omega_t=\Omega(c_t), \qquad
o_{t+1}=o_t \cup \{(c_t,\omega_t)\},
\end{equation}
which is appended to the interaction history and used in the next reasoning step. The loop continues until the model outputs the final answer or reaches a predefined interaction budget.

\noindent\textbf{Thought \(r_t\).}
In our framework, \(r_t\) corresponds to the key decision and reasoning stages. Starting from the raw input, the model first determines whether the current context already provides a standardized facial view; if not, it invokes a face normalization tool by default. Once a valid facial view is available, the thought further decides whether additional local evidence is needed, i.e., whether and where to zoom in. After new observations are returned, the thought also determines how the available evidence should be processed through \emph{Global Analysis}, \emph{Local Analysis}, or their combination.

\noindent\textbf{Action \(c_t\).}
The action space \(\mathcal{C}\) contains two vision tools. (1) \textit{Face Detection-Alignment}: Performs face detection and landmark-based alignment based on the face analysis toolkit InsightFace~\cite{ren2023pbidr,guo2021sample,gecer2021ostec}. (2) \textit{Zoom-In}: Crops and magnifies a selected facial region to expose local expression cues that may be difficult to resolve from the holistic view alone.

\noindent\textbf{Observation \(\omega_t\).}
The observation \(\omega_t\) denotes the feedback returned by tool execution. For \textit{Face Detection-Alignment}, \(\omega_t\) consists of the aligned face crop when face detection succeeds, together with the bounding-box coordinates \((x_1, y_1, x_2, y_2)\) of four semantically meaningful facial regions, namely \emph{forehead--eyebrow}, \emph{eye--periorbital}, \emph{nose}, and \emph{mouth--chin}. These regions serve as candidate local evidence for subsequent inspection. For the \textit{Zoom-In} tool, \(\omega_t\) consists of magnified local ROI images obtained by cropping the input according to the selected region coordinates.


\subsection{Training Data Curation}

\noindent\textbf{Synthetic Process.}
We curate data from four FER datasets: AffectNet~\cite{mollahosseini2017affectnet}, FERPlus~\cite{barsoum2016training}, RAF-DB~\cite{li2017reliable}, and SFEW2.0~\cite{zhang2024generalizable}, covering eight emotion categories. As shown in the Stage~1 block of Figure~\ref{fig:framework_overview} (b), for each image, we first run the two tools used during agent interaction offline to obtain an aligned face and the associated ROI crops. We then use Qwen3VL-235B-A22B-Instruct~\cite{bai2025qwen3} to generate dense facial descriptions for both the raw image and the processed views. Based on these descriptions, we leverage another strong language model, gpt-oss-120b~\cite{agarwal2025gpt}, together with injected FACS knowledge to extract a pseudo AU set for each sample. Finally, we synthesize multi-turn trajectories that contain tool calls, intermediate reasoning, AU predictions, and final emotion answers. To match the three execution modes of ActFER, the synthetic trajectories explicitly include (i) full trajectories with detection and adaptive zoom, (ii) simplified trajectories with detection but no zoom, and (iii) degraded trajectories that fall back to holistic reasoning when no reliable face can be detected.

\begin{figure}[t]
    \centering
    \vspace{-0.5cm}
    \includegraphics[width=0.97\linewidth]{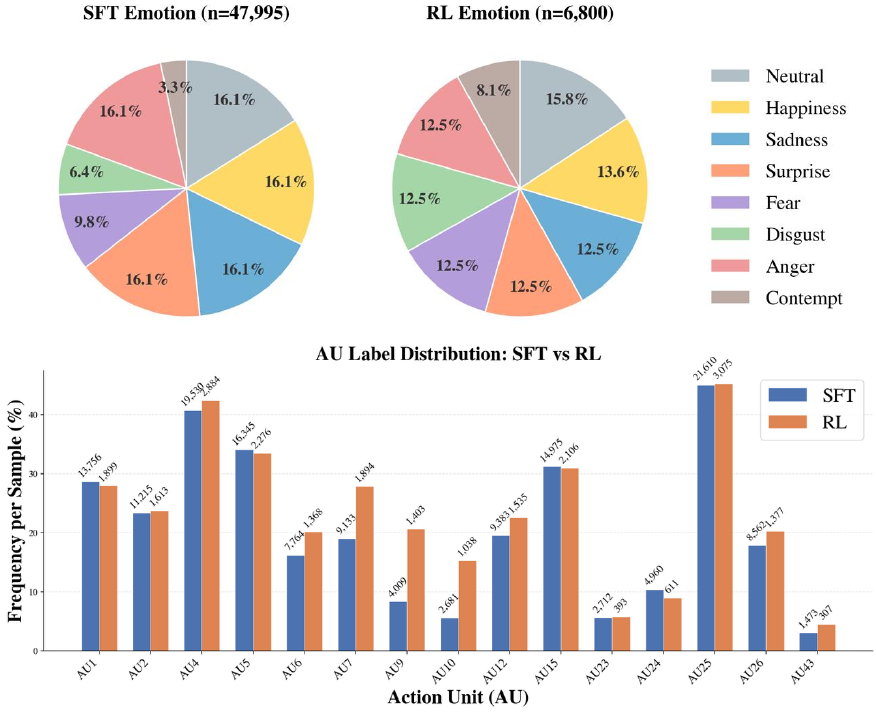}
    \caption{Statistics of the curated training data.}
    \Description{A single-column combined figure summarizing the curated training data. It includes the emotion distributions of the SFT and RL subsets and a bar chart comparing AU frequencies between the two splits. The RL subset is more balanced across emotions, especially for low-resource categories, while preserving similar AU coverage.}
    \label{fig:data_stats}
    \vspace{-0.2cm}
\end{figure}

\noindent\textbf{Sampling Strategy.}
After trajectory construction, we apply quality filtering and class-aware sampling to form disjoint SFT and RL subsets, with 48K samples for SFT and 6.8K for RL. The SFT subset preserves a broader natural label distribution, allowing the model to learn the general tool-use protocol and response format. The RL subset is deliberately re-balanced toward harder low-resource emotions. Figure~\ref{fig:data_stats} visualizes the resulting data statistics.

\subsection{Supervised Fine-Tuning}
The first training stage teaches the model to follow a structured tool-use protocol and produce the expected Visual CoT format. We perform full-parameter fine-tuning of Qwen3VL-4B~\cite{bai2025qwen3} on the multi-turn trajectories. Given a demonstration trajectory $\tau^{\star} = (a_1, \dots, a_{|\tau^{\star}|})$, the SFT objective is the standard autoregressive loss over the full action-and-response sequence:
\begin{equation}
\mathcal{L}_{\text{SFT}} = -\sum_{t=1}^{|\tau^{\star}|} \log \pi_{\theta}(a_t \mid x, a_{<t}).
\end{equation}
This objective jointly supervises tool calls, intermediate reasoning, AU prediction, and final emotion output, providing a stable cold start for subsequent UC-GRPO optimization.

\subsection{UC-GRPO: Utility-Calibrated GRPO}
The second stage uses Utility-Calibrated GRPO (UC-GRPO), our task-adaptive extension of GRPO~\cite{shao2024deepseekmath} for agentic FER.
Naive reward design has two limitations. First, a reward based only on the final emotion is too sparse and coarse to reveal whether local zooming improves the perception of fine-grained facial movements. Second, a fixed reward or penalty for tool use cannot distinguish two opposite errors: missing a beneficial local inspection and performing an unnecessary one. As shown in the Stage~2 block of Figure~\ref{fig:framework_overview} (b), we decompose the reward into three terms: (1) an AU-grounded dense task reward $R_{\text{acc}}$ that combines emotion correctness with AU evidence, (2) a tool reward $R_{\text{tool}}$ that integrates rule-based structural constraints and utility calibration for local inspection, and (3) a bounded quality reward $R_{\text{qual}}$ that discourages low-quality outputs. The overall reward is
\begin{equation}
\mathcal{R}(\tau_i) = \lambda R_{\text{acc}}(\tau_i) + (1-\lambda) R_{\text{tool}}(\tau_i) + R_{\text{qual}}(\tau_i),
\label{eq:overall_reward}
\end{equation}
where $\tau_i$ denotes the $i$-th rollout in the current query group.

\noindent\textbf{AU-Grounded Dense Task Reward.}
The direct effect of local inspection is usually not an immediate change in the final emotion label. In fact, it first changes how clearly the model can perceive local facial movements. We therefore densify task supervision by coupling emotion correctness with AU-set quality. For a predicted AU set $\hat{A}$ and target set $A^{\star}$, we use the instance-level AU-set F1:
\begin{equation}
F_{1}^{\text{AU}}(\hat{A}, A^{\star}) = \frac{2|\hat{A}\cap A^{\star}|}{|\hat{A}| + |A^{\star}|}.
\label{eq:au_f1}
\end{equation}
The resulting task reward is
\begin{equation}
R_{\text{acc}}=
\begin{cases}
 w_y + w_{au} F_{1}^{\text{AU}}(\hat{A}, A^{\star}), & \hat{y}=y^{\star},\\
 r_{\text{wrong}} + \frac{1}{2} \, w_{au} F_{1}^{\text{AU}}(\hat{A}, A^{\star}), & \hat{y}\neq y^{\star},
\end{cases}
\label{eq:acc_reward}
\end{equation}
where $y^{\star}$ is the target emotion label, $w_y$ and $w_{au}$ form a convex decomposition of the task reward ($w_y=1-w_{au}$), and $r_{\text{wrong}}<0$ penalizes incorrect emotion prediction. Even when the final emotion is wrong, we retain half of the AU-based term so that partially correct local evidence is not discarded completely. This shifts supervision from asking only whether the final class is correct to asking whether the trajectory captures the facial evidence relevant to emotion recognition. 

\begin{algorithm}[t]
\caption{ActFER training with UC-GRPO}
\label{alg:facer}
\begin{algorithmic}[1]
\REQUIRE input image $x$, group size $G$, maximum turns $T$
\FOR{each rollout $\tau_i$, $i=1,\dots,G$}
    \STATE Initialize the context with $x$
    \FOR{$t=1$ to $T$}
        \STATE Sample an action $u_{i,t} \sim \pi_\theta(\cdot \mid o_{i,t})$
        \IF{$u_{i,t}$ is a tool invocation}
            \STATE Execute the tool and update the context
        \ELSE
            \STATE Produce AU prediction $\hat{A}_i$ and emotion prediction $\hat{y}_i$; terminate rollout
        \ENDIF
    \ENDFOR
    \STATE Compute task reward $R_{\text{acc}}(\tau_i)$ and record zoom usage $z_i$
\ENDFOR
\STATE Partition rollouts into groups: $\mathcal{Z}^{+}(q)$ and $\mathcal{Z}^{-}(q)$
\STATE Compute the utility gap $\Delta(q)$
\FOR{each rollout $\tau_i$}
    \STATE Compute utility reward $R_{\text{util}}(\tau_i)$ by Eq.~\eqref{eq:util_cascade}
    \STATE Compute total reward $\mathcal{R}(\tau_i)$ by Eq.~\eqref{eq:overall_reward}
\ENDFOR
\STATE Normalize rewards within the group and update $\theta$ with GRPO
\STATE Update emotion-wise EMA statistics
\end{algorithmic}
\end{algorithm}
\noindent\textbf{FSM-Constrained Structural Reward.}
We decompose tool-side supervision $R_{\text{tool}}$ into structural validity $R_{\text{fsm}}$ and inspection utility $R_{\text{util}}$. The structural term $R_{\text{fsm}}$ serves as an executable scaffold, which checks the response format, penalizes parse errors, illegal tool order, repeated or excessive tool calls, and rules out impossible behaviors such as zooming after failed face detection. This keeps utility learning grounded in legal and executable trajectories.

\noindent\textbf{Utility-Calibrated Inspection Reward.}
The inspection term $R_{\text{util}}$ determines whether zooming is worthwhile for the current sample and how strongly the resulting decision should be rewarded or penalized. It combines a short-horizon query-level contrastive estimate with a long-horizon emotion-level calibration, and falls back to a symmetric performance-gated signal. 

\begin{table*}[t]
    \caption{Main emotion-recognition results. The left block reports benchmark-level F1 (\%), overall Acc (\%) and F1 (\%) on FERBench; the right block reports per-emotion F1 (\%). * InternVL3.5-4B~\cite{wang2025internvl3}, Qwen3VL-4B~\cite{bai2025qwen3}, EmoLA~\cite{li2024facial}, and ExpLLM~\cite{Lan2025ExpLLM} are reproduced from official checkpoints, with evaluation strictly following the FERBench protocol. Results of other open-source baselines are taken from UniFER~\cite{zhang2025rethinking}. Best results are \textbf{boldfaced}; second best are \underline{underlined}.}
    \label{tab:main_emotion_results}
    \centering
    \small
    \setlength{\tabcolsep}{3.0pt}
    \resizebox{\textwidth}{!}{%
    \begin{tabular}{lcccccc|cccccccc}
        \toprule
        \multirow{2}{*}{Model} & \multicolumn{4}{c}{Benchmark F1} & \multicolumn{2}{c|}{Overall} & \multicolumn{8}{c}{Per-emotion F1} \\
        \cmidrule(r){2-5}\cmidrule(r){6-7}\cmidrule(l){8-15}
        & RAF-DB & FERPlus & AffectNet & SFEW2.0 & Acc & F1 & Neutral & Happiness & Sad & Surprise & Fear & Disgust & Anger & Contempt \\
        \midrule
        \multicolumn{15}{l}{\textit{General-purpose MLLMs}} \\
        InternVL3.5-4B*~\cite{wang2025internvl3} & 60.69 & 37.87 & 44.51 & 33.96 & 50.40 & 43.67 & 55.56 & 81.89 & 53.47 & 55.06 & 17.44 & 35.59 & 46.68 & 3.70 \\
        Qwen3VL-4B*~\cite{bai2025qwen3} & 56.13 & 47.37 & 36.12 & 38.57 & 55.27 & 44.03 & 60.12 & 81.93 & 54.01 & 56.06 & 22.97 & 23.47 & 50.00 & 3.65 \\
        Qwen2.5VL-7B~\cite{bai2025qwen25vltechnicalreport} & 50.26 & 46.25 & 26.97 & 36.22 & 53.78 & 35.34 & 54.60 & 67.29 & 53.91 & 52.70 & 3.28 & 22.26 & 46.97 & 0.37 \\
        InternVL3-8B~\cite{zhu2025internvl3} & 52.69 & 43.34 & 39.33 & 37.27 & 60.59 & 44.54 & 65.08 & 85.56 & 61.57 & 40.02 & 41.35 & 38.31 & 54.41 & 14.55 \\
        LLaVA-Next-34B~\cite{liu2024improved} & 60.56 & 48.43 & 34.92 & 37.76 & 61.20 & 44.36 & 60.62 & \underline{86.11} & 64.00 & 53.79 & 3.01 & 35.79 & 51.22 & 0.36 \\
        Gemini-2.5-Flash~\cite{google2025gemini25flashcard} & 55.60 & 44.95 & 45.38 & 37.20 & 61.55 & 45.47 & 62.25 & 81.53 & 59.69 & 63.43 & 42.52 & 41.82 & 48.86 & 9.14 \\
        Gemini-2.5-Pro~\cite{google2025gemini25procard} & 50.95 & 39.78 & 43.11 & 36.33 & 57.17 & 44.29 & 47.43 & 81.22 & 57.78 & 63.16 & 42.42 & 39.66 & 46.44 & \underline{20.49} \\
        \midrule
        \multicolumn{15}{l}{\textit{MLLM-based FER methods}} \\
        EmoLA* <7B>~\cite{li2024facial} & 53.46 & 45.29 & 35.12 & 37.91 & 62.75 & 43.19 & 55.13 & 77.47 & 47.05 & 48.67 & 28.70 & 33.50 & 38.37 & 16.62 \\
        ExpLLM* <7B>~\cite{Lan2025ExpLLM} & \textbf{84.80} & 54.37 & \underline{46.86} & 43.49 & \underline{69.33} & \underline{57.26} & 68.52 & 84.78 & \underline{71.23} & \underline{69.58} & \underline{54.79} & \underline{46.83} & 62.36 & 0.00 \\
        UniFER-7B RL~\cite{zhang2025rethinking} & 81.30 & \underline{58.55} & 44.53 & 39.70 & 68.84 & 55.32 & \underline{72.80} & 84.32 & \textbf{72.46} & 51.31 & 51.57 & 41.68 & \textbf{65.08} & 3.31 \\
        \midrule
        \rowcolor{lightgray} \textbf{ActFER-SFT <4B>} & 74.87 & 51.94 & 45.53 & \underline{46.61} & 65.70 & 53.37 & 70.33 & 84.60 & 65.44 & 64.78 & 41.03 & 41.54 & 58.14 & 1.11 \\
        \rowcolor{lightgray} \textbf{ActFER <4B>} & \underline{82.72} & \textbf{59.92} & \textbf{57.66} & \textbf{51.13} & \textbf{73.89} & \textbf{67.45} & \textbf{79.28} & \textbf{89.34} & 70.39 & \textbf{70.90} & \textbf{60.00} & \textbf{54.71} & \underline{63.97} & \textbf{51.00} \\
        \bottomrule
    \end{tabular}}
\end{table*}

\paragraph{Query-Conditional Contrastive Utility Estimation.}
Once dense task reward is available, UC-GRPO estimates the utility of local inspection by comparing rollouts within the same query group.

For a query $q$, let $\{\tau_1, \ldots, \tau_G\}$ be the $G$ rollouts sampled for the same image. We first exclude trajectories with face-detection failure and partition the remaining rollouts into those using \texttt{zoom\_in} and those not using:
\begin{equation}
\mathcal{Z}^{+}(q)=\{i \mid \texttt{zoom\_in}\in \tau_i\}, \quad
\mathcal{Z}^{-}(q)=\{i \mid \texttt{zoom\_in}\notin \tau_i\}.
\end{equation}
When both subsets are non-empty, we estimate the query-level utility of local inspection directly from the task reward:
\begin{equation}
\Delta(q)=\frac{1}{|\mathcal{Z}^{+}(q)|}\sum_{i\in\mathcal{Z}^{+}(q)} R_{\text{acc}}(\tau_i)-
\frac{1}{|\mathcal{Z}^{-}(q)|}\sum_{i\in\mathcal{Z}^{-}(q)} R_{\text{acc}}(\tau_i),
\label{eq:delta_reward}
\end{equation}
where $\Delta(q)>0$ indicates that zooming is beneficial for the current sample, while $\Delta(q)<0$ indicates harmful. Using $R_{\text{acc}}$ keeps utility estimation aligned with emotion accuracy and AU quality.

\paragraph{EMA Calibration.}
Although $\Delta(q)$ provides a sample-aware estimate of zoom utility, it can still be noisy because it is computed from a finite rollout group. Moreover, different emotions may rely on local evidence to different degrees. To capture such category-level tendencies while smoothing noisy sample-level signals, we maintain an exponential moving average (EMA) of query-level utility for each emotion category:
\begin{equation}
\bar{\Delta}_e^{(n)} = \rho \, \hat{\Delta}_e^{(n)} + (1-\rho) \, \bar{\Delta}_e^{(n-1)},
\label{eq:ema_update}
\end{equation}
where $\hat{\Delta}_e^{(n)}$ is the mean of all valid $\Delta(q)$ whose ground-truth emotion is $e$ at training step $n$. The resulting $\bar{\Delta}_e^{(n)}$ summarizes the historical tendency of zoom to help or hurt emotion $e$. We then convert $\bar{\Delta}_e$ into two bounded modulation factors, $\phi_e^{\text{lazy}}$ and $\phi_e^{\text{unnec}}$, using a monotone sigmoid-based map: emotions with historically positive zoom utility penalize missed zoom opportunities more strongly, whereas emotions with historically negative utility penalize unnecessary zooming more strongly. To avoid early instability, this calibration is activated only after sufficient statistics have been accumulated. Before that, both factors are set to 1. 

For a rollout $\tau_i$ from query $q$, let $z_i\in\{0,1\}$ indicate whether the trajectory invokes \texttt{zoom\_in}. Given $z_i$ and the query-level utility gap $\Delta(q)$, we define the adaptive utility reward as:
\begin{equation}
\label{eq:adaptive_zoom}
\begin{aligned}
R_{\text{util}}^{\text{adapt}}(z_i,\Delta(q),e)
&=
\begin{cases}
 r_{\text{pos}}, & \substack{|\Delta(q)|<\varepsilon,\\
                    \Delta(q)\ge \varepsilon \land z_i=1,\ \text{or}\\
                    \Delta(q)\le -\varepsilon \land z_i=0,}\\
 -h(|\Delta(q)|)\phi_e^{\text{lazy}}, & \Delta(q)\ge \varepsilon \ \text{and}\ z_i=0,\\
-h(|\Delta(q)|)\phi_e^{\text{unnec}}, & \Delta(q)\le -\varepsilon \ \text{and}\ z_i=1.
\end{cases}
\end{aligned}
\end{equation}
Here, $r_{\text{pos}}>0$ denotes a fixed positive reward for utility-consistent decisions and neutral cases. The margin $\varepsilon$ defines a neutral interval around zero, within which zoom and no-zoom are treated as approximately equivalent. The factor $\phi_e^{\text{lazy}}$ is applied only when a beneficial zoom is missed, while $\phi_e^{\text{unnec}}$ is applied only when an unnecessary zoom is taken. The penalty magnitude $h(|\Delta(q)|)$ is a bounded increasing function implemented with a tanh-shaped curve and shared fixed constants.


\paragraph{Symmetric Fallback Reward.}
In practice, some rollout groups do not contain sufficient zoom diversity, for example when all rollouts choose zoom or all skip it. In such cases, $\Delta(q)$ cannot be estimated reliably. Rather than fabricating pseudo-utility, we use a symmetric performance-gated fallback:
\begin{equation}
R_{\text{util}}^{\text{fb}}(s_i)=
\begin{cases}
 r_{\text{pos}}, & s_i \geq s_{\text{high}},\\
 r_{\text{neg}}, & \text{otherwise},
\end{cases}
\label{eq:fallback_reward}
\end{equation}
We define the coarse task-performance score as
\begin{equation*}
s_i=\max\!\left\{\mathbb{I}[\hat{y}_i=y^{\star}],F_{1}^{\text{AU}}(\hat{A}_i,A^{\star})\right\}.
\end{equation*}
This fallback does not encode a fixed zoom preference; it only prevents instability when direct contrastive evidence is unavailable.

The final utility term for rollout $\tau_i$ is therefore
\begin{equation}
R_{\text{util}}(\tau_i)=
\begin{cases}
R_{\text{util}}^{\text{adapt}}(z_i,\Delta(q),e), & \Delta(q)\text{ is defined},\\
R_{\text{util}}^{\text{fb}}(s_i), & \text{otherwise},
\end{cases}
\label{eq:util_cascade}
\end{equation}
and the overall tool reward is
\begin{equation}
R_{\text{tool}}(\tau_i) = R_{\text{fsm}}(\tau_i) + R_{\text{util}}(\tau_i).
\end{equation}
In implementation, utility calibration is computed in two passes within each GRPO update. We first obtain rollout-level task rewards and tool-use metadata, then estimate $\Delta(q)$ at the query-group level and inject either the adaptive or fallback reward. EMA statistics are updated after the current step, while reward construction for the current step uses the pre-update EMA state.

\noindent\textbf{Quality Regularization and GRPO Update.}
In addition to task and tool rewards, we use a bounded regularizer $R_{\text{qual}}$ to mildly penalize low-quality outputs such as severe formatting errors, repetition, and language mixing. After computing the total reward in Eq.~\eqref{eq:overall_reward}, GRPO normalizes rewards within each query group as
\begin{equation}
\widehat{\mathrm{Adv}}_i = \frac{\mathcal{R}(\tau_i) - \mu_q}{\sigma_q},
\end{equation}
where $\mu_q$ and $\sigma_q$ are the mean and standard deviation of rewards within the rollout group for query $q$. This group-relative ranking is well suited to utility learning, because it encourages the policy to prefer better evidence-acquisition strategies among alternative trajectories for the same face. Algorithm~\ref{alg:facer} summarizes the overall interaction and optimization procedure at the query-group level.



\section{Experiments}
\label{sec:experiments}

\subsection{Implementation Details}
We build our training framework on VERL~\cite{sheng2024hybridflow}, an open-source RL library that supports agent rollouts with tool use. Unless otherwise noted, all variants use the same Qwen3VL-4B~\cite{bai2025qwen3} backbone, the same mixed training split, and the same interaction budget. In SFT, we fine-tune all LLM linear layers on 48K tool-grounded trajectories while freezing the vision encoder. We use AdamW with a base learning rate of $1 \times 10^{-5}$ for the LLM and $2 \times 10^{-5}$ for the vision-language projector, and weight decay 0.01. We adopt an effective batch size of 16, a maximum sequence length of 8192 tokens, and train for 2 epochs. In RL, we start from the SFT checkpoint and optimize with UC-GRPO on the 6.8K single-turn prompts for policy rollouts. We again use AdamW, reduce the learning rate to $2 \times 10^{-6}$, and use a batch size of 32. The interaction horizon is capped at $T \leq 4$, GRPO uses $G=5$ rollouts per query, the maximum prompt length is 8192 tokens, and the maximum response length is 4096 tokens. To stabilize training, we apply KL regularization with coefficient 0.1, set the entropy coefficient to 0.01, and normalize advantages by the within-group standard deviation. RL runs for 1 epoch. All experiments are conducted on four NVIDIA A800 (80GB) GPUs.

\begin{figure}[t]
    \centering
    \includegraphics[width=0.98\linewidth]{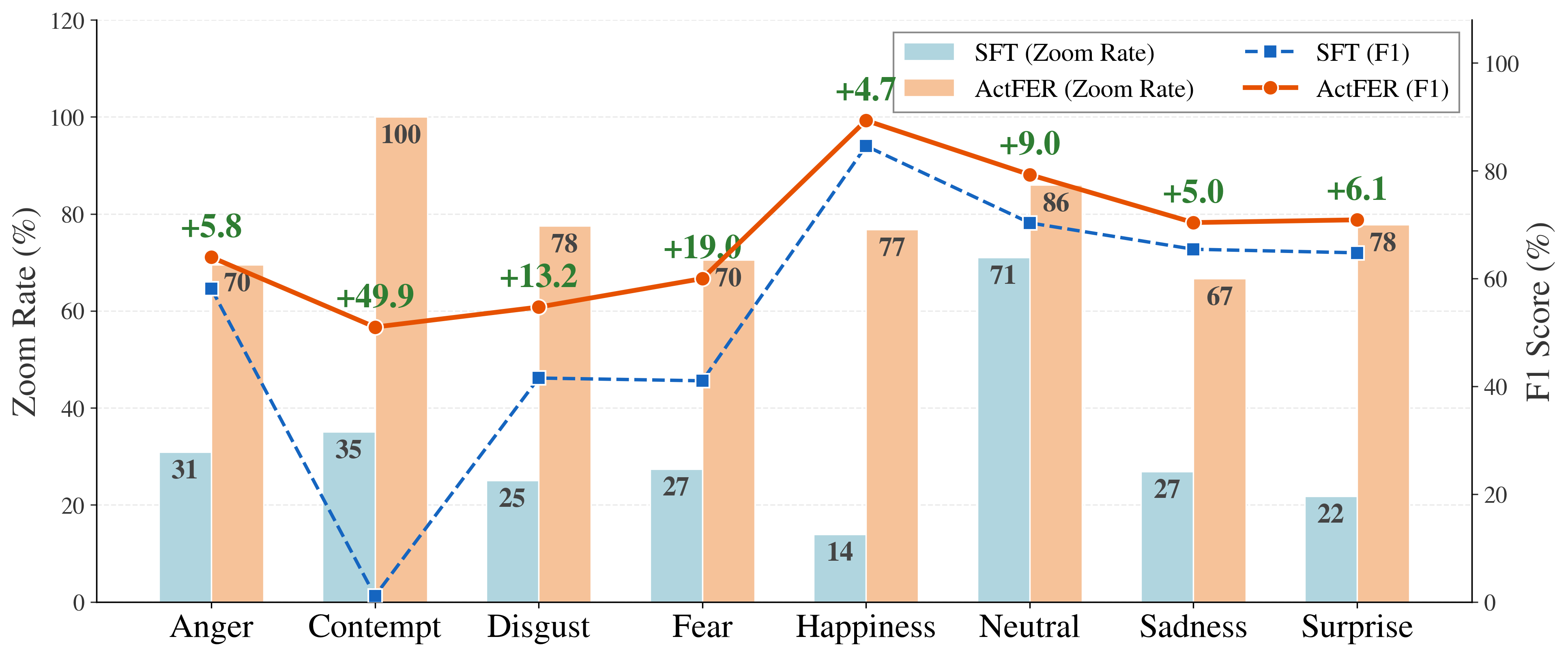}
    \caption{Emotion-wise comparison between ActFER-SFT and ActFER on FERBench. Bars denote the proportion of invoking zoom, and lines denote the per-emotion F1.} 
    \Description{A two-column comparison plot over eight emotions. For each emotion, two bars show the zoom ratio of ActFER-SFT and Full ActFER, and two lines show their corresponding F1 scores. Full ActFER increases zoom usage across all emotions, with the strongest increase and largest F1 improvement on contempt.}
    \label{fig:emotion_zoom_vs_f1}
\end{figure}

\begin{table}[t]
    \caption{Comparison of zero-shot AU F1 (\%) on DISFA under the standard 8-AU protocol. The table lists the F1 score for each individual AU together with the average F1 (Avg F1) over all 8 AUs. * InternVL3.5-4B~\cite{wang2025internvl3} and Qwen3VL-4B~\cite{bai2025qwen3} are reproduced on DISFA using their official checkpoints. 
    }
    \label{tab:au_results}
    \scriptsize
    \centering
    \setlength{\tabcolsep}{3.0pt}
    \resizebox{\columnwidth}{!}{%
    \begin{tabular}{@{}lccccccccc@{}}
        \toprule
        Method & 1 & 2 & 4 & 6 & 9 & 12 & 25 & 26 & Avg. \\
        \midrule
        Qwen2.5VL-7B~\cite{bai2025qwen25vltechnicalreport} & 20.8 & 19.2 & 23.0 & 17.9 & 17.1 & 26.3 & 30.9 & 21.6 & 22.1 \\
        Qwen3VL-4B*~\cite{bai2025qwen3} & 24.7 & 27.1 & \underline{76.7} & 3.1 & 0.0 & \underline{73.7} & 80.7 & 18.6 & 38.1 \\
        InternVL3.5-4B*~\cite{wang2025internvl3} & \underline{36.9} & \textbf{40.7} & 71.8 & 2.1 & \underline{35.3} & 67.6 & 67.2 & 4.6 & 40.8 \\
        FEALLM <7B>~\cite{hu2025feallm} & \underline{36.9} & 29.7 & 70.4 & 30.2 & -- & 54.5 & 51.9 & \underline{24.7} & 42.6 \\
        \midrule
        \rowcolor{lightgray} \textbf{ActFER-SFT <4B>} & 30.2 & 30.8 & 70.2 & \underline{51.4} & 44.1 & 67.7 & \underline{82.0} & 12.5 & \underline{48.6} \\
        \rowcolor{lightgray} \textbf{ActFER <4B>} & \textbf{37.1} & \underline{38.2} & \textbf{77.3} & \textbf{58.2} & \textbf{52.3} & \textbf{78.0} & \textbf{92.4} & \textbf{32.0} & \textbf{58.2} \\
        \bottomrule
    \end{tabular}}
\end{table}

\begin{table*}[t]
    \caption{UC-GRPO ablation of ActFER. The configuration columns indicate whether the AU-grounded dense reward $\mathcal{R}_{acc}$ is enabled and which components are included in the tool-side reward $\mathcal{R}_{tool}$. The FERBench block reports emotion Accuracy/F1 and the zoom ratio (i.e., the percentage of samples that invoke zoom); the DISFA block reports average AU F1 and the zoom ratio. For \textit{Zoom-biased RL}, the entire utility reward $\mathcal{R}_{util}$ is disabled, including both the adaptive branch and the symmetric fallback, and is replaced by a constant zoom-preference bonus to bias the policy toward using zoom.}
    \vspace{-0.2cm}
    \label{tab:component_ablation}
    \small
    \centering
    \setlength{\tabcolsep}{4.2pt}
    \begin{tabular}{@{}lccccccccc@{}}
        \toprule
        \multirow{2}{*}{Variant} & {$\mathcal{R}_{acc}$} & \multicolumn{3}{c}{$\mathcal{R}_{tool}$} & \multicolumn{3}{c}{FERBench} & \multicolumn{2}{c}{DISFA} \\
        \cmidrule(lr){2-2} \cmidrule(lr){3-5} \cmidrule(lr){6-8} \cmidrule(lr){9-10}
        & \shortstack{AU-grounded dense reward} & \shortstack{Contrastive utility} & \shortstack{Emotion-EMA} & \shortstack{Symmetric fallback} & Acc & F1 & \shortstack{Zoom} & \shortstack{Avg.F1} & \shortstack{Zoom} \\
        \midrule
        ActFER-SFT & -- & -- & -- & -- & 65.70 & 53.37 & 32.6 & 48.6 & 8.4 \\
        \midrule
        Emotion-only RL & \xmark{} & \xmark{} & -- & \cmark{} & 69.14 & 61.49 & 0.0 & 47.8 & 0.0 \\
        AU-grounded RL & \cmark{} & \xmark{} & -- & \cmark{} & 68.80 & 62.72 & 0.0 & 50.1 & 0.0 \\
        Zoom-biased RL & \cmark{} & \xmark{} & -- & \xmark{} & 71.86 & 62.22 & 100.0 & 52.0 & 100.0 \\
        \shortstack[l]{w/o Emotion-EMA} & \cmark{} & \cmark{} & \xmark{} & \cmark{} & 70.01 & 65.34 & 90.1 & 54.2 & 100.0 \\
        \rowcolor{lightgray} \textbf{Full ActFER} & \textbf{\cmark{}} & \textbf{\cmark{}} & \textbf{\cmark{}} & \textbf{\cmark{}} & \textbf{73.89} & \textbf{67.45} & 77.5 & \textbf{58.2} & 91.9 \\
        \bottomrule
    \end{tabular}
    \vspace{-0.2cm}
\end{table*}


\subsection{Comparison on Emotion Recognition} Following the FERBench protocol introduced in UniFER~\cite{zhang2025rethinking}, we evaluate ActFER on the test sets of AffectNet~\cite{mollahosseini2017affectnet}, RAF-DB~\cite{li2017reliable}, FERPlus~\cite{barsoum2016training}, and SFEW2.0~\cite{zhang2024generalizable} without per-dataset tuning. The comparison includes general-purpose MLLMs, MLLM-based FER methods under the same protocol, and our own variants. Table~\ref{tab:main_emotion_results} summarizes both benchmark-level and per-emotion results. ActFER achieves the best overall performance, reaching 73.89 Accuracy and 67.45 macro-F1. It surpasses the strongest general-purpose MLLM, Gemini-2.5-Flash~\cite{google2025gemini25flashcard}, by +12.34 Acc and +21.98 F1, and also outperforms the strongest FER-specific baselines, demonstrating the effectiveness of agentic local inspection.
The per-emotion results show clear category-dependent gains. ActFER achieves new best results on neutral, happiness, surprise, fear, disgust, and contempt, with the largest improvement on contempt, where F1 reaches 51.00 versus the previous best of 20.49. Figure~\ref{fig:emotion_zoom_vs_f1} further shows that zoom usage varies substantially across emotions, and the resulting F1 gains are far from uniform. This suggests that the benefit of ActFER does not come from simply zooming more often, but from learning when local inspection is most useful. The especially large gain on contempt is consistent with this interpretation: as a subtle and low-resource category, contempt is more likely to benefit from selective inspection of weak mouth-region cues, while also receiving greater emphasis from the RL split on harder emotions. We therefore attribute this result to utility-calibrated, category-aware inspection rather than a generic increase in zoom frequency. 

\begin{figure}[t]
    \centering
    \includegraphics[width=\linewidth]{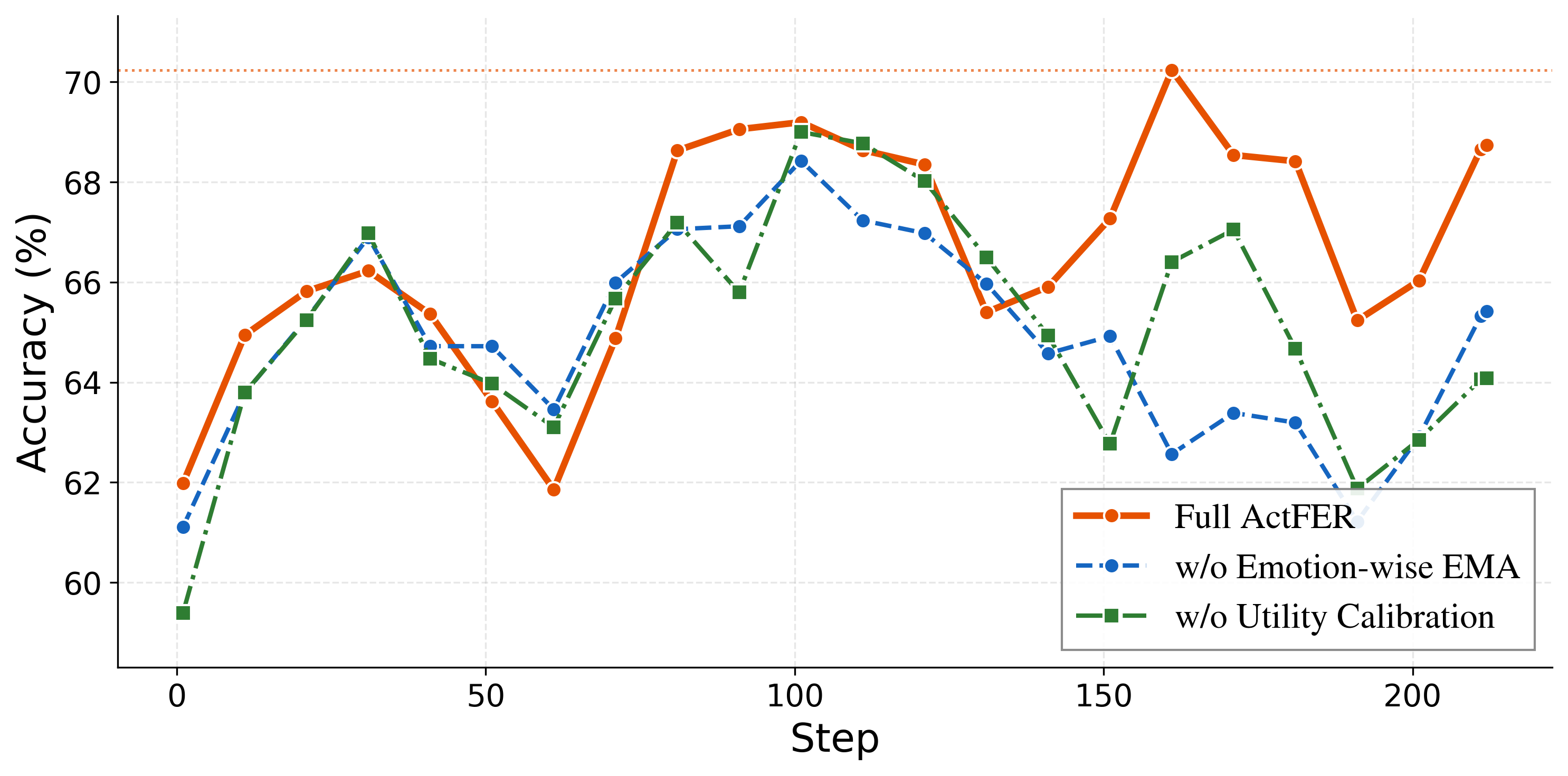}
    \vspace{-0.4cm}
    \caption{Training-time emotion accuracy for three utility-calibration variants. Full ActFER attains the best late-stage accuracy plateau, while removing EMA or utility calibration results in policy collapse during the later stage of training.}
    \Description{A single line chart comparing training-time emotion accuracy for Full ActFER, a variant without emotion-wise EMA, and a variant without utility calibration. Full ActFER reaches the highest late-stage plateau, the w/o EMA variant oscillates more, and the w/o utility variant converges lower.}
    \label{fig:training_dynamics}
    \vspace{-0.3cm}
\end{figure}

\subsection{Comparison on Zero-shot AU Detection}
We further evaluate zero-shot AU transfer on DISFA~\cite{mavadati2013disfa} under the common 8-AU protocol, with the results summarized in Table~\ref{tab:au_results}. Considering that DISFA is annotated from continuous video frames, we perform test-time subsampling for efficiency, keeping at most five frames with the same AU label for each subject.
ActFER achieves the best zero-shot transfer result, reaching 58.2 average AU F1 without DISFA training. This is +20.1 over the base Qwen3VL-4B and +15.6 over FEALLM, although FEALLM is explicitly designed to strengthen local-detail perception. Relative to FEALLM, ActFER gains especially on AU6, AU12, and AU25. These AUs rely heavily on localized eye, nose, and especially mouth-region cues, showing that active local magnification is more effective than static single-pass local modeling for structured facial-movement perception. Overall, ActFER delivers state-of-the-art zero-shot AU performance under the no-DISFA-tuning setting.


\begin{figure*}[t]
    \centering
    \begin{minipage}[c]{0.63\textwidth}
        \centering
        \includegraphics[width=\linewidth]{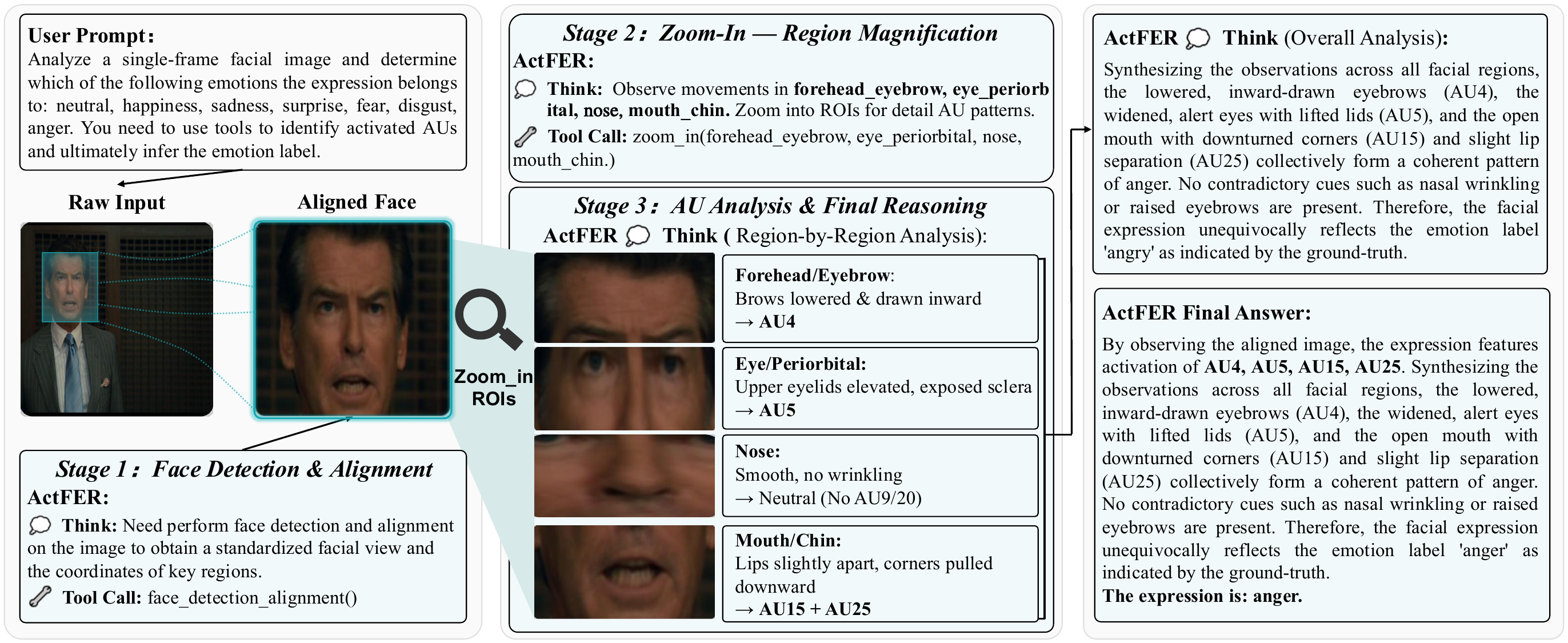}
        \small (a) End-to-end reasoning on an anger example.
    \end{minipage}\hfill
    \begin{minipage}[c]{0.35\textwidth}
        \centering
        \includegraphics[width=\linewidth]{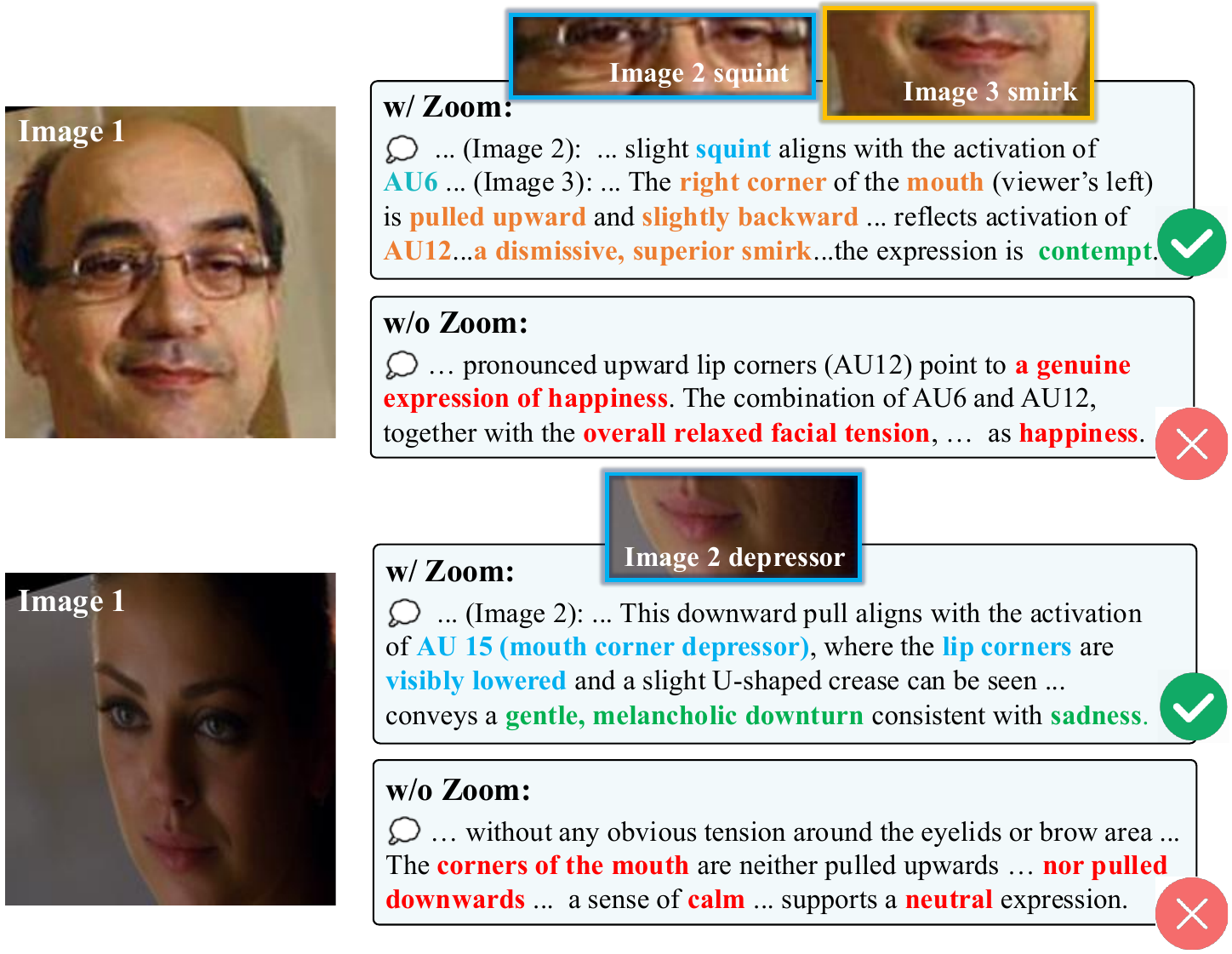}
        \small (b) Zoom/no-zoom comparisons.
    \end{minipage}
    \caption{Qualitative analyses of ActFER. (a) The model aligns the face, invokes \texttt{zoom\_in} on task-relevant regions, analyzes local AU evidence, and predicts anger. (b) On subtle contempt and sadness examples, adaptive zoom reveals weak local cues that the same model misses without zoom.}
    \Description{Two qualitative analyses. Panel a shows an end-to-end ActFER trajectory for an anger example: the system aligns a face, zooms into four facial regions, identifies AU4, AU5, AU15, and AU25, and predicts anger. Panel b compares adaptive zoom with no zoom. For contempt, zoom reveals a slight squint and asymmetric upward mouth-corner pull and corrects a happiness prediction; for sadness, zoom reveals the downward mouth-corner pattern of AU15 and corrects a neutral prediction.}
    \label{fig:qualitative_cases}
    \label{fig:zoom_vs_nozoom_cases}
\end{figure*}

\subsection{Ablation Study}

\noindent \textbf{Component Ablation of ActFER.} Table~\ref{tab:component_ablation} reports the UC-GRPO ablation on FERBench emotion recognition and DISFA zero-shot AU transfer. Four findings emerge. First, all RL variants outperform ActFER-SFT, showing that the gains arise from reward-driven optimization rather than tool grounding alone. Second, adding $R_{\text{acc}}$ improves both macro-F1 and AU transfer, confirming the value of AU-grounded supervision. However, Emotion-only RL and AU-grounded RL both collapse to a no-zoom policy, indicating that end-task supervision alone cannot teach when local inspection is useful. Third, \textit{Zoom-biased RL} outperforms these no-zoom variants but remains inferior to contrastive utility modeling, suggesting that the key is sample-specific utility estimation rather than zoom itself. Finally, emotion-wise EMA further improves both benchmarks while reducing zoom usage relative to w/o Emotion-EMA, demonstrating better calibration rather than simple tool suppression. Overall, ActFER benefits from task-aligned rewards, utility-aware tool selection, and stable long-horizon calibration.

\noindent \textbf{Zoom-Policy Analysis across Training Strategies.} Table~\ref{tab:component_ablation} reveals two zoom-policy collapse modes. Without utility modeling, Emotion-only RL and AU-grounded RL never zoom on either benchmark, showing that symmetric tool rewards and stronger task supervision are insufficient to induce useful local inspection. Conversely, contrastive utility without emotion-wise EMA causes aggressive over-zooming, likely because noisy sample-level utility estimates push the policy toward an extreme. Full ActFER avoids both failure modes. As shown in Figure~\ref{fig:training_dynamics}, w/o Emotion-EMA exhibits larger training oscillations, whereas the full model converges more smoothly to a higher accuracy plateau. Thus, emotion-wise EMA stabilizes utility learning and supports a balanced inspection policy. Full ActFER retains substantial zoom usage (77.5\% on FERBench and 91.9\% on DISFA) while achieving the best emotion recognition and zero-shot AU transfer performance.

\subsection{Qualitative Results}
Figure~\ref{fig:qualitative_cases}(a) shows a successful ActFER trajectory on a challenging anger sample. The agent aligns the face, selectively inspects ROIs, identifies AU evidence, and correctly predicts the emotion, illustrating its global-to-local, AU-grounded reasoning process.

Figure~\ref{fig:zoom_vs_nozoom_cases}(b) compares ActFER with adaptive zoom against the same model without zoom on two subtle expressions. For \textit{contempt}, zoom exposes a slight squint and an asymmetric upward mouth-corner pull, distinguishing contempt from a coarse happiness interpretation. For \textit{sadness}, zoom reveals the downward mouth-corner pattern associated with AU15, whereas the no-zoom model misses this weak cue and predicts neutral. These cases show how adaptive zoom can turn ambiguous global impressions into more reliable predictions when decisive evidence is weak and localized.

\section{Conclusion}
We presented ActFER, an agentic framework for MLLM-based facial expression recognition. By combining lightweight perceptual tools, FACS-grounded reasoning, and Utility-Calibrated GRPO, ActFER reformulates FER as utility-aware local inspection followed by structured affective inference, rather than passive single-pass reasoning over fixed inputs. Beyond improving emotion recognition, ActFER promotes a more structured inference process: preparing analyzable facial evidence, inspecting informative local regions, inferring facial movements, and then reasoning to the emotion label. Accordingly, our experiments evaluate not only final predictions but also the quality of the learned evidence preparation and inspection process.

\begin{acks}
This work was supported in part by a grant from the National Natural Science
Foundation of China (No.~62406264) and a grant from the Anhui Natural Science
Foundation (No.~2508085ZD006).
\end{acks}

\bibliographystyle{ACM-Reference-Format}
\balance
\bibliography{references}

\end{document}